\documentclass[letterpaper]{article} 
\usepackage[preprint]{paperstyle}  
\usepackage[hyphens]{url}  
\usepackage{graphicx} 
\usepackage{natbib}  
\usepackage{caption} 
\usepackage{algorithm}
\usepackage{algorithmic}
\usepackage{colortbl}
\usepackage{xcolor}
\definecolor{best}{RGB}{255, 0, 0}   
\definecolor{second}{RGB}{0, 0, 255} 
\definecolor{rowgray}{gray}{0.90}

\newcommand{\cgred}{\cellcolor{best!10}}
\newcommand{\cgblue}{\cellcolor{second!10}}
\usepackage{amsmath}
\usepackage{amssymb}
\usepackage{mathtools}
\usepackage{amsthm}
\usepackage{amsfonts}
\usepackage{threeparttable}
\usepackage{multirow}
\usepackage{subcaption}
\usepackage{newfloat}
\usepackage{listings}
\usepackage{bm}
\DeclareCaptionStyle{ruled}{labelfont=normalfont,labelsep=colon,strut=off} 
\floatstyle{ruled}
\newfloat{listing}{tb}{lst}{}
\floatname{listing}{Listing}
\usepackage{booktabs}

\title{Rarity-Aware Discrete Diffusion with Spatially Consistent Decoding for Photo-Realistic Image Super-Resolution}
\author{
Ao Li\textsuperscript{\rm 1},
Yapeng Du\textsuperscript{\rm 1},
Yi Xin\textsuperscript{\rm 2},
Lei Zhu\textsuperscript{\rm 3},
Le Zhang\textsuperscript{\rm 1},
Guangtao Zhai\textsuperscript{\rm 4},
Ce Zhu\textsuperscript{\rm 1}\corresponding,
Xiaohong Liu\textsuperscript{\rm 4,5}\corresponding
}
\affiliations{
\textsuperscript{\rm 1}University of Electronic Science and Technology of China\\
\textsuperscript{\rm 2}Nanjing University\\
\textsuperscript{\rm 3}The Hong Kong University of Science and Technology (Guangzhou)\\
\textsuperscript{\rm 4}Shanghai Jiao Tong University\\
\textsuperscript{\rm 5}Shanghai Innovation Institute\\
}

\begin{document}

\maketitle

\begin{abstract}
  Continuous diffusion models have become the dominant paradigm for photo-realistic image Super-Resolution (SR), but they typically formulate reconstruction as continuous signal-level denoising and incorporate semantic priors through external conditioning modules. This makes it less direct to exploit the unified token-based scaling paradigm of modern multimodal models. Autoregressive models provide a more native semantic representation by modeling images as discrete visual tokens, yet their causal decoding is inefficient for high-resolution reconstruction. Discrete diffusion offers a promising middle ground by enabling non-causal, parallel prediction over visual tokens. However, directly adapting discrete diffusion to SR remains non-trivial due to two task-specific challenges: (1) the long-tailed distribution of visual tokens, which under-represents rare but perceptually critical textures; and (2) spatially inconsistent parallel decoding, which may introduce isolated artifacts. To address these issues, we propose DiMOO-SR, a rarity-aware multimodal discrete diffusion framework for photo-realistic image SR. During training, Inverse Frequency Sampling (IFS) prioritizes under-represented but information-rich tokens. During inference, Spatial Consistency Ranking (SCR) refines token confidence using local neighborhood agreement to improve structural coherence. Extensive experiments on widely used real-world SR benchmarks demonstrate that DiMOO-SR achieves competitive perceptual quality with only a few parallel decoding steps, highlighting the potential of discrete diffusion for generative image super-resolution. The code will be released upon publication.
\end{abstract}

\section{Introduction}
\label{sec:introduction}
Photo-realistic image Super-Resolution (SR) strives to recover intricate high-frequency details that align with both low-level textures and high-level semantics. Continuous diffusion models have become the dominant paradigm for photo-realistic image SR~\cite{saharia2022image, rombach2022high}. However, they typically formulate reconstruction as continuous signal-level denoising, and semantic priors are often introduced through external conditioning modules. This makes the interaction between low-level reconstruction and high-level semantic modeling less native than unified token-based generation. To introduce the necessary ``understanding'' into the reconstruction loop, many contemporary frameworks attempt to bridge this gap by integrating external multi-modal priors. Typically, these methods graft pre-trained semantic encoders (e.g., CLIP~\cite{radford2021learning}) onto the denoising backbone to provide high-level guidance. Although effective, this coupling introduces an additional interface between semantic features and the reconstruction backbone. As a result, semantic guidance is often injected through external conditioning rather than modeled within a unified token sequence.

A more unified alternative emerged with the Autoregressive (AR) multi-modal paradigm, exemplified by PURE~\cite{Wei_2025_ICCV}. By quantizing images into discrete tokens, PURE leverages the contextual reasoning of Large Language Models (LLMs) to natively perceive and understand scene content before reconstruction. However, PURE's reliance on causal sequence generation creates a severe efficiency bottleneck: tokens are predicted strictly one-by-one. Generating a high-resolution image requires over a thousand sequential Transformer passes, making it prohibitively slow for practical deployment.

This brings us to a critical question: \textbf{\textit{Can we achieve native semantic understanding in a non-causal, efficient manner?}} The emergence of Discrete Diffusion Models (DDMs) for global synthesis~\cite{xin2025lumina, nie2025large} suggests a promising ``third path''. By employing parallel masked prediction, DDMs maintain the unified token-based understanding of AR models while restoring the entire image in just a dozen iterations. Recent discrete SR efforts such as ITER~\cite{chen2024iter} show that token-space diffusion is a viable alternative to continuous denoising. However, the potential of large multimodal DDMs for semantic-aware SR remains underexplored, and our investigation reveals that a vanilla adaptation is suboptimal due to two domain-specific design bottlenecks. 

During training, general DDMs are commonly trained with uniform masking, which fails to account for the information density of SR tasks. We observe that visual codebooks are heavily long-tailed: common low-frequency tokens, such as sky and walls, dominate the distribution, while rare texture tokens are statistically under-represented. As a result, a uniform strategy may allocate excessive capacity to trivial background reconstruction and insufficient supervision to perceptually important textures.

On the inference stage, parallel DDMs predict masked tokens independently based on pointwise confidence. During the rapid iterative refinement of SR, this lack of a structural consensus mechanism allows isolated errors—tokens that are confident but spatially inconsistent with their neighbors, resulting in artifacts that erode structural integrity.

To overcome these obstacles, we propose DiMOO-SR, a specialized discrete diffusion framework for photo-realistic image SR. Our contributions are summarized as follows.
\begin{enumerate}
  \item We identify token rarity imbalance and spatially inconsistent decoding as two practical bottlenecks when adapting discrete diffusion to image super-resolution.
  \item We introduce Inverse Frequency Sampling (IFS), which reallocates training supervision toward rare but information-rich visual tokens.
  \item We propose Spatial Consistency Ranking (SCR), a parameter-free decoding strategy that refines confidence ranking through local neighborhood agreement.
  \item Experiments on RealSR and DRealSR show that DiMOO-SR improves perceptual and distributional metrics under a controlled token-based generative SR setting.
\end{enumerate}

\section{Related Works}
\label{sec:related_works}
\subsection{Image Super-Resolution}
Single image super-resolution (SR) has evolved through distinct phases. Early regression-based methods~\cite{liang2021swinir,chen_tpami_2025,li2025exploring,zhou2023srformer} utilized MSE-based objectives to achieve high PSNR. However, they notoriously suffer from the ``regression-to-mean'' problem, producing over-smoothed results lacking high-frequency details. To improve perceptual quality, GAN-based approaches~\cite{wang2021real,zhang2021designing,wang2018esrgan} introduced adversarial training, yet they often generate hallucinatory artifacts. Most recently, Continuous Diffusion Models (CDMs) have dominated the field. Methods such as StableSR~\cite{wang2024exploiting}, PASD~\cite{yang2024pixel}, and SeeSR~\cite{wu2024seesr} leverage priors from frozen text-to-image models to synthesize realistic textures. Many continuous diffusion SR methods require iterative denoising, and recent accelerated variants such as OSEDiff~\cite{wu2024one} and SinSR~\cite{wang2024sinsr} have significantly reduced the number of sampling steps. Our goal is therefore not to claim a universal speed advantage over all continuous diffusion methods, but to explore a complementary discrete token-generation paradigm that naturally aligns with multimodal token modeling.

\subsection{Discrete Visual Generation}
Discrete generative models quantize images into visual tokens via VQ-VAE~\cite{van2017neural} or VQ-GAN~\cite{esser2021taming}, enabling the application of Transformer architectures. This field is bifurcated into two streams: Autoregressive (AR) models and Discrete Diffusion Models (DDMs). Autoregressive models predict tokens sequentially, typically in a raster-scan or multi-scale order. Recent advances have successfully adapted this paradigm to SR: PURE~\cite{Wei_2025_ICCV} integrates multimodal understanding into an AR transformer. However, the intrinsic sequential nature of AR models limits their inference speed and restricts global context modeling during the early generation stages. Conversely, DDMs like MaskGIT~\cite{chang2022maskgit}, Lumina-mGPT~\cite{liu2026lumina,xin2025luminamgpt}, and Lumina-DiMOO~\cite{xin2025lumina} employ a parallel iterative decoding strategy. For real-world SR, ITER~\cite{chen2024iter} performs distortion removal followed by token-space discrete diffusion with token evaluation and refinement. Different from ITER, our goal is to unlock large multimodal DDM priors for semantic-aware SR. We show that standard DDM treatment of visual tokens remains suboptimal because rare detail tokens are under-trained and parallel decoding lacks spatial consensus.

\section{Empirical Analysis}
\label{sec:empirical}

\subsection{The Statistical Imbalance of Visual Codebooks}
\label{subsec:imbalance}
\begin{figure}[!t]
  \centering
  \includegraphics[width=\linewidth]{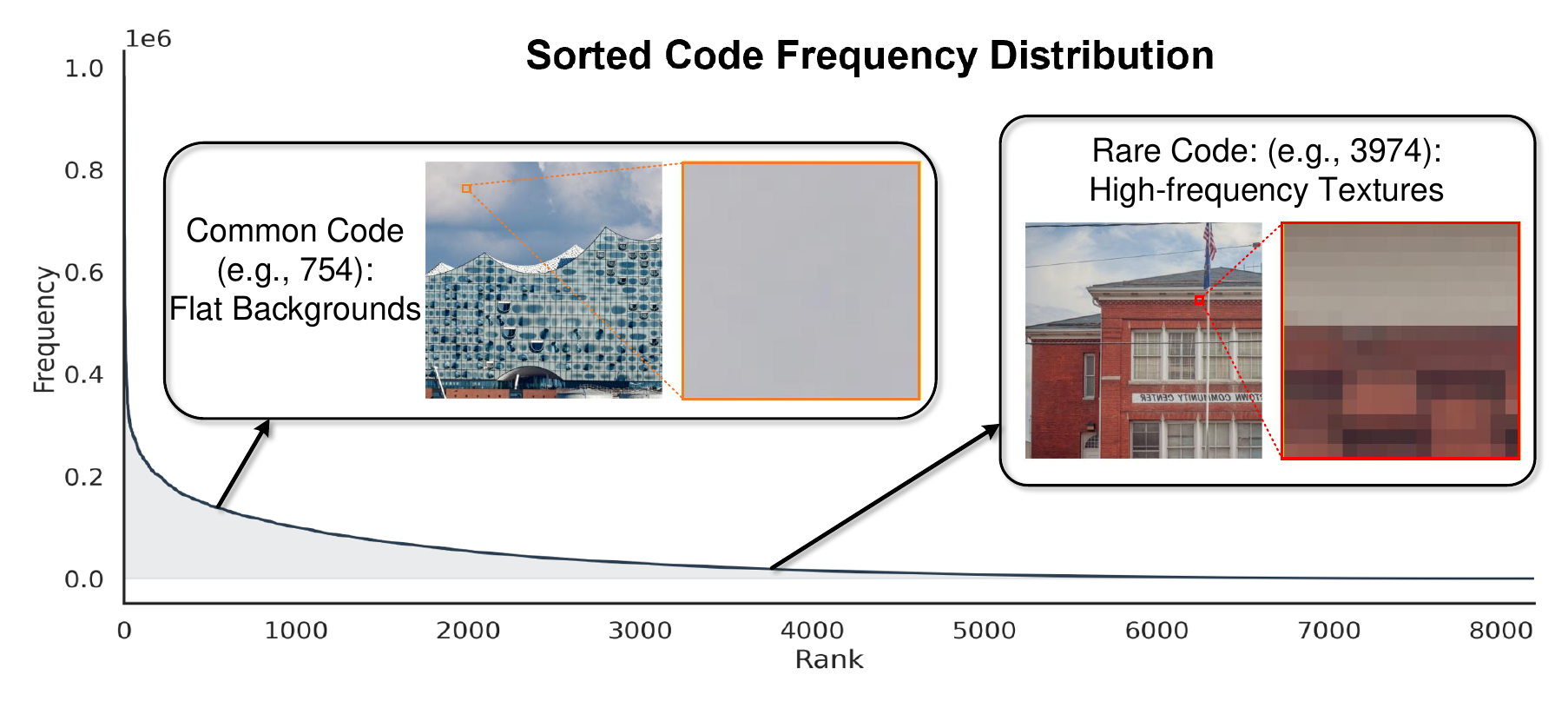}
  \caption{
    Correlation between rarity and semantic density. The codebook follows a long-tailed distribution. Common codes (e.g., 754) correspond to flat backgrounds, while rare codes (e.g., 3974) encode high-frequency textures. Random masking predominantly samples the former, leading to inefficient fine-tuning for super-resolution.
    }
  \label{fig:token_analysis}
\end{figure}

Our investigation is grounded in the discrete latent space of the Lumina-DiMOO~\cite{xin2025lumina}. Let $\mathbf{X} \in \mathbb{R}^{H \times W \times 3}$ denote a high-resolution image patch. The VQ-GAN encoder $\mathcal{E}$ maps $\mathbf{X}$ to a continuous latent map, which is then quantized element-wise using a learnable codebook $\mathcal{Z} = \{z_k\}_{k=1}^K$ with vocabulary size $K=8192$. The quantization process for a spatial position $(i, j)$ is defined as:
\begin{equation}
    q(z_{i,j}) = \mathop{\arg\min}_{z_k \in \mathcal{Z}} \| \mathcal{E}(\mathbf{X})_{i,j} - z_k \|_2.
\end{equation}
This operation results in a discrete token map $\mathbf{M} \in \{1, \dots, K\}^{h \times w}$, where $h=H/16$ and $w=W/16$. To characterize the usage distribution of $\mathcal{Z}$, we performed a global statistical sweep across our dataset $\mathcal{D}$ comprising approximately $3 \times 10^5$ image patches. We define the empirical marginal probability of the $k$-th code index as $p(k) = \frac{1}{N} \sum_{\mathbf{M} \in \mathcal{D}} \sum_{i=1}^{hw} \mathbb{I}(M_i = k)$, where $N$ is the total number of tokens across the dataset. The resulting distribution, as visualized in Figure \ref{fig:token_analysis}, exhibits a severe long-tailed characteristic. Specifically, the distribution reveals extreme sparsity: the top 40\% of indices account for over 90\% of the total usage, while the remaining tokens form a long tail. Our data reveals that a small ``Head'' fraction of the codebook accounts for the vast majority of latent activations, while the ``Tail'' consists of thousands of rare tokens that appear sporadically.
\begin{figure*}[!t]
  \centering
  \includegraphics[width=\linewidth]{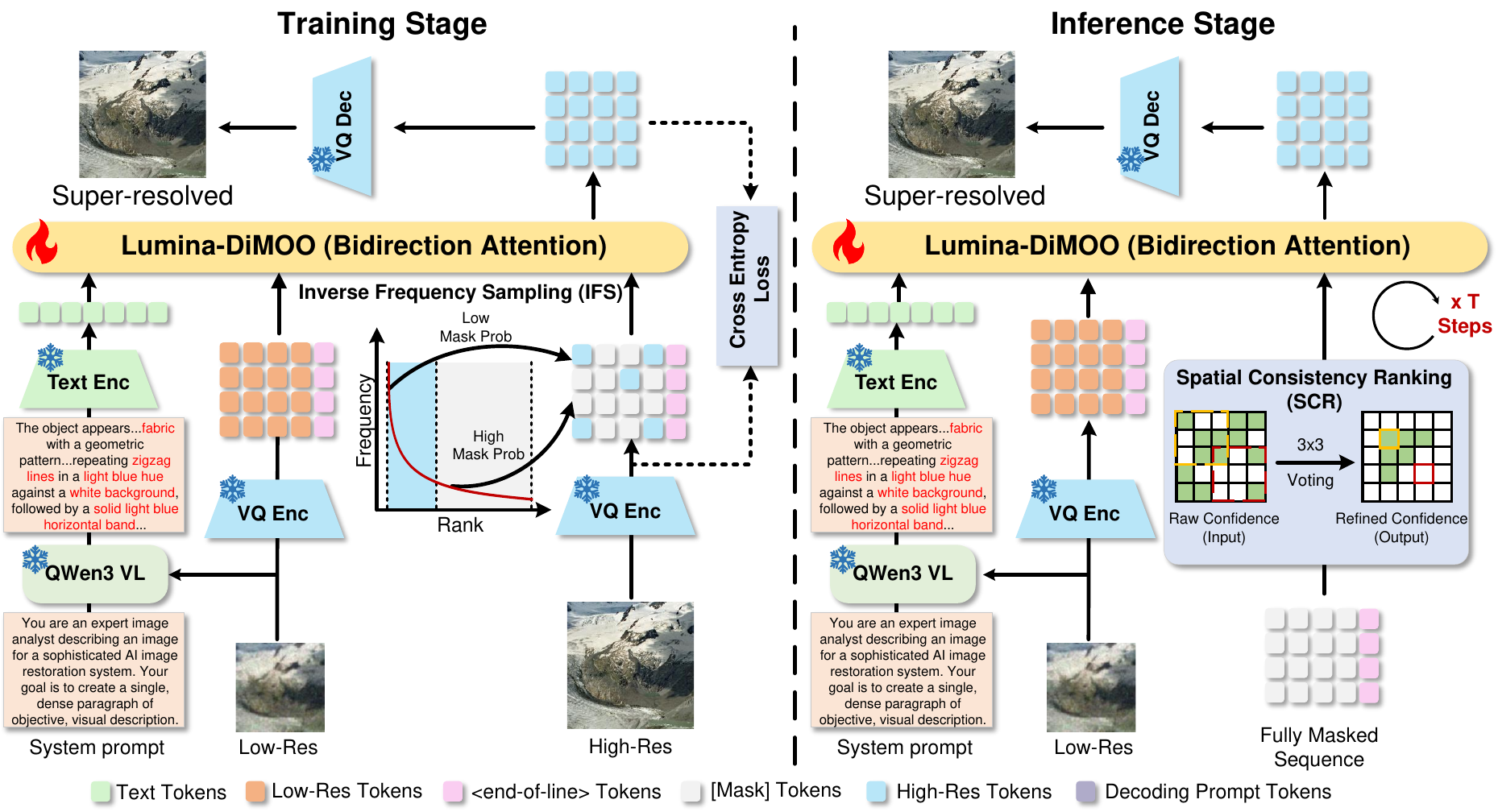}
  \caption{
  Overview of the proposed framework. 
  Left (Training): We replace uniform masking with Inverse Frequency Sampling (IFS), assigning higher masking probabilities to rare, texture-rich tokens (visualized by the density curve). Right (Inference): We introduce Spatial Consistency Ranking (SCR), a parameter-free module that refines token confidence using anchor-aware local consensus to mitigate isolated artifacts.
  }
  \label{fig:framework}
\end{figure*}
\subsection{Information-Theoretic Interpretation of Rarity}
To understand the semantic implications of this imbalance, we analyze the tokens through the lens of Information Theory. We quantify the information content of a specific token index $k$ using its \textit{Self-Information}:
\begin{equation}
    I(k) = - \log p(k).
\end{equation}
This formulation implies a fundamental dichotomy: frequent ``Head'' tokens possess low self-information ($I(k) \to 0$), signifying redundancy, whereas rare ``Tail'' tokens carry high self-information, encoding the unique signals in the data. We corroborate this theoretical insight with a reverse activation mapping protocol. By projecting the receptive fields of specific tokens back to the pixel space, we observe a distinct semantic separation, as shown in Figure~\ref{fig:token_analysis}. Tokens with low $I(k)$ almost exclusively map to low-frequency, homogeneous regions (e.g., clear skies). In contrast, tokens with high $I(k)$ consistently correspond to high-frequency structural details (e.g., fabric textures, complex edges). This confirms that in the context of VQ-GAN, \textit{statistical rarity is a proxy for local structural complexity}.

\textbf{The Limitations of Random Masking.} This heavy-tailed distribution exposes a fundamental bottleneck in methods that employ uniform random masking for SR. Under uniform sampling, the training objective is statistically dominated by the abundant ``Head'' tokens. Consequently, the model biases its optimization towards reconstructing smooth backgrounds, largely neglecting the recovery of fine textures. This imbalance leads to artifacts analogous to ``regression-to-mean'', where the model generates safe, blurry predictions rather than sharp, perceptually realistic details.

\section{Methodology}
\label{sec:methodology}
Motivated by the statistical insights from our empirical analysis, we propose a unified framework for photo-realistic super-resolution based on the Lumina-DiMOO~\cite{xin2025lumina} architecture. As illustrated in Figure~\ref{fig:framework}, our approach integrates two key innovations: Inverse Frequency Sampling (IFS) during training to rebalance the learning of texture-rich tokens, and Spatial Consistency Ranking (SCR) during inference to ensure structural coherence.

\subsection{Model Architecture}
Our backbone is a transformer with bidirectional attention, enabling global context modeling. We first discretize the $512 \times 512$ high-resolution image into a $32 \times 32$ token map $\mathbf{x}_{\text{HR}} \in \mathcal{V}^{1024}$ using a pre-trained VQ-GAN encoder~\cite{xin2025lumina}. To incorporate multimodal conditioning, we employ a Qwen3-VL encoder~\cite{qwen3-vl-instruct-hf} to extract text embeddings from the input caption, which are then tokenized into a sequence $\mathbf{x}_{\text{text}} \in \mathcal{V}^{L}$. Following the Lumina-DiMOO paradigm, we concatenate $\mathbf{x}_{\text{HR}}$, $\mathbf{x}_{\text{text}}$, and the condition tokens $\mathbf{x}_{\text{LR}}$ derived from the input image along the sequence dimension, forming a unified token stream $\mathbf{x} = [\mathbf{x}_{\text{text}}; \mathbf{x}_{\text{LR}};\mathbf{x}_{\text{HR}}]$. This concatenated sequence is directly fed into the transformer, allowing seamless multimodal interaction through bidirectional self-attention without explicit cross-attention modules.

\subsection{Training: Inverse Frequency Sampling}
\label{subsec:ifs}
To rectify the statistical bias identified above, we introduce Inverse Frequency Sampling (IFS). Let $\bm{x}=\{x_1,\cdots,x_N\}$ denote the flattened sequence of visual tokens, where $i\in\{1,\cdots,N\}$ represents the spatial index. Standard uniform masking strategies assign equal probability $P_{\text{uni}}(i) = 1/N$ to all positions, regardless of their semantic density. IFS reformulates this as an importance sampling process governed by the information content of each token. The IFS probability distribution is defined as:
\begin{equation}
  P_{\text{IFS}}(i) = \frac{\exp(\alpha \cdot I(x_i))}{\sum_{j=1}^N \exp(\alpha \cdot I(x_j))},
\end{equation}
where $\alpha$ serves as the information temperature. As $\alpha$ increases, the model performs increasingly aggressive hard negative mining, allocating its limited capacity strictly toward reconstructing texture-rich details rather than trivial backgrounds.

\textbf{Hybrid Sampling Strategy.} 
Exclusively training on rare tokens with high masking ratios can destabilize optimization, as the model may lack sufficient context to infer high-information details, leading to ``context collapse''. To balance \textit{structural generation} (from scratch) and \textit{textural refinement}, we design a hybrid masking strategy governed by a Bernoulli parameter $p=0.5$. Let $\gamma$ denote the mask ratio (percentage of tokens masked). During training, we alternate between two pathways:
\begin{equation}
    \mathbf{m} \sim 
    \begin{cases} 
    \text{Sample}(P_{\text{uni}}, \gamma \in [0, 1]) & \text{with prob } 0.5 \\
    \text{Sample}(P_{\text{IFS}}, \gamma \in [0, 0.6]) & \text{with prob } 0.5.
    \end{cases}
\end{equation}
In the first pathway, the model acts as a \textit{Global Structure Branch}, applying standard uniform masking across the full spectrum of generation with $\gamma \sim \mathcal{U}(0, 1)$. This ensures the model learns to construct consistent global layouts from an empty canvas, providing the necessary robustness for the early steps of inference. In the second pathway, the model functions as a \textit{Hard Refinement Branch}, applying IFS with a truncated ratio $\gamma \sim \mathcal{U}(0, 0.6)$. By limiting the mask ratio to a maximum of $60\%$, we ensure that at least $40\%$ of the visual context is preserved. This constraint prevents context collapse, providing necessary structural cues that allow the model to focus on synthesizing plausible fine textures from sufficient context.

\textbf{Defensive mixture view.}
IFS can be interpreted as reweighting supervision toward rare tokens, which may increase gradient variance when the token distribution is long-tailed. Hybrid sampling forms a \emph{defensive mixture} that ensures every token has nonzero probability of being sampled through the uniform component. Concretely, let $q_{\text{uni}}(x)$ and $q_{\text{IFS}}(x)$ denote the sampling probabilities induced by the two masking rules, and let
\begin{equation}
q_{\text{hybrid}}(x) = p\,q_{\text{uni}}(x) + (1-p)\,q_{\text{IFS}}(x).
\end{equation}
The effective importance weight of the IFS component within the hybrid distribution is
\begin{equation}
\frac{q_{\text{IFS}}(x)}{q_{\text{hybrid}}(x)} \,=\, \frac{q_{\text{IFS}}(x)}{p\,q_{\text{uni}}(x) + (1-p)\,q_{\text{IFS}}(x)} \,\le\, \frac{1}{1-p},
\end{equation}
which is bounded by $2$ when $p=0.5$. Under the mild assumption that individual per-token gradients are bounded, this bound prevents arbitrarily large importance weights and provides a principled explanation for why the hybrid strategy is typically more stable than pure IFS.

\begin{table*}[!ht]
  \centering
  \caption{Cross-paradigm quantitative comparison. 
  We evaluate DiMOO-SR against \textbf{continuous diffusion models} and \textbf{token-based generative models}. Ours is listed in both sections.
  \colorbox{red!10}{\textcolor{red}{Red}} and \colorbox{blue!10}{\textcolor{blue}{blue}} indicate the best and second-best performance within each block.
  }
  \label{tab:cmp_sota}
  \footnotesize
  \setlength{\tabcolsep}{9.3pt}
  \begin{threeparttable}
  \begin{tabular}{c|l|ccccccc}
    \toprule
    Dataset & Method & PSNR$\uparrow$ & SSIM$\uparrow$ & LPIPS$\downarrow$ & DISTS$\downarrow$ & FID$\downarrow$ & MANIQA$\uparrow$ & NIQE$\downarrow$ \\
    \midrule
    \multicolumn{9}{c}{\textbf{\textit{Comparison I: vs. Continuous Diffusion Models}}} \\
    \midrule
    \multirow{4}{*}{\textit{DRealSR}} 
      & DiffBIR  & 26.71 & 0.6571 & 0.4557 & 0.2748 & 166.79 & 0.5930 & \cgblue\textcolor{second}{6.3124} \\

      & PASD & 27.36 & 0.7073 & 0.3760 & 0.2531 & 156.13 & \cgred\textcolor{best}{0.6169} & \cgred\textcolor{best}{5.5474} \\

      & SeeSR & \cgred\textcolor{best}{28.17} & \cgred\textcolor{best}{0.7691} & \cgred\textcolor{best}{0.3189} & \cgblue\textcolor{second}{0.2315} & \cgblue\textcolor{second}{147.39} & \cgblue\textcolor{second}{0.6042} & 6.3967 \\

      & \textbf{Ours} & \cgblue\textcolor{second}{27.78} & \cgblue\textcolor{second}{0.7442} & \cgblue\textcolor{second}{0.3431} &\cgred \textcolor{best}{0.2259} & \cgred\textcolor{best}{143.45} & 0.5409 & 6.4552\\

    \midrule

    \multirow{4}{*}{\textit{RealSR}} 
      & DiffBIR  & 24.75 & 0.6567 & 0.3636 & 0.2312 & 128.99 & 0.6246 & 5.5346 \\

      & PASD  & \cgred\textcolor{best}{25.21} & 0.6798 & 0.3380 & 0.2260 & \cgblue\textcolor{second}{124.29} & \cgred\textcolor{best}{0.6487} & 5.4137 \\

      & SeeSR  & \cgblue\textcolor{second}{25.18} & \cgred\textcolor{best}{0.7216} & \cgred\textcolor{best}{0.3009} & \cgblue\textcolor{second}{0.2223} & {125.55} & \cgblue\textcolor{second}{0.6442} & \cgblue\textcolor{second}{5.4081} \\

      & \textbf{Ours} & 24.21 & \cgblue\textcolor{second}{0.6887} & \cgblue\textcolor{second}{0.3053} & \cgred\textcolor{best}{0.2110} & \cgred\textcolor{best}{123.78} & 0.6034 & \cgred\textcolor{best}{5.3636} \\

    \midrule
    \midrule
    \multicolumn{9}{c}{\textbf{\textit{Comparison II: vs. Token-Based Generative Models}}} \\
    \midrule
    \multirow{5}{*}{\textit{DRealSR}} 
      & PURE    & 26.08 & 0.6823 & {0.3853} & {0.2487} & 165.01 & \cgred\textcolor{best}{0.5683} & \cgblue\textcolor{second}{6.6328} \\

      & ITER & 25.17 & 0.7385 & 0.3451 & \cgblue\textcolor{second}{0.2294} & 180.83 & 0.4222 & 7.3730 \\

      & Lumina-DiMOO  & \cgred\textcolor{best}{28.44}  &	\cgred\textcolor{best}{0.7771}	& \cgblue\textcolor{second}{0.3450} 	& \cgblue\textcolor{second}{0.2383} 	& \cgblue\textcolor{second}{153.56}	&   0.5000  & 	7.5381 \\

      & \textbf{Ours} & \cgblue\textcolor{second}{27.78} & \cgblue\textcolor{second}{0.7442} & \cgred\textcolor{best}{0.3431} &\cgred \textcolor{best}{0.2259} & \cgred\textcolor{best}{143.45} & \cgblue\textcolor{second}{0.5409} & \cgred\textcolor{best}{6.4552}\\

    \midrule

    \multirow{5}{*}{\textit{RealSR}} 
      & PURE & 23.47 & 0.6545 & 0.3400 & 0.2336 & 136.05 & \cgred \textcolor{best}{0.6052} & \cgblue\textcolor{second}{5.6703}\\

      & ITER & 24.14 & \cgred\textcolor{best}{0.7201} & \cgblue\textcolor{second}{0.3072} & \cgblue\textcolor{second}{0.2192} & 163.84 & 0.5027 & 6.8220 \\

      & Lumina-DiMOO & \cgred\textcolor{best}{24.86}  &	\cgblue \textcolor{second}{0.7114}	& 0.3085 	& 0.2233 	& \cgblue \textcolor{second}{133.54} 	&   0.5649  & 	5.9649   \\

      & \textbf{Ours} & \cgblue\textcolor{second}{24.21} & 0.6887 & \cgred\textcolor{best}{0.3053} & \cgred\textcolor{best}{0.2110} & \cgred\textcolor{best}{123.78} & \cgblue\textcolor{second}{0.6034} & \cgred\textcolor{best}{5.3636} \\
    \bottomrule
  \end{tabular}
\end{threeparttable}
\end{table*}

\subsection{Inference: Spatial Consistency Ranking}
Although IFS improves the learning of rare visual tokens, standard parallel decoding still ranks masked tokens independently according to point-wise confidence. Such independent ranking may retain locally inconsistent predictions, especially when visually isolated tokens receive high confidence before sufficient spatial context is established. To mitigate this issue, we introduce Spatial Consistency Ranking (SCR), which refines the confidence ranking used for re-masking during parallel decoding.

At each decoding step, the model samples a candidate token $\hat{x}_i$ for each currently masked visual position. We define the raw confidence as the probability assigned to the sampled candidate:
\begin{equation}
c_i = p_\theta(\hat{x}_i \mid x_{\mathrm{obs}}, x_{\mathrm{LR}}, x_{\mathrm{text}}).
\end{equation}
This definition is consistent with stochastic token sampling, where a sampled token is considered reliable only when the model assigns sufficient probability to that specific candidate.

SCR incorporates spatial context through an anchor-aware confidence field. Let $C^{(t)}$ denote the confidence field over the compact visual-token grid at decoding step $t$. For visual tokens that have already been retained in previous iterations, we set $C^{(t)}_j=1$ and treat them as reliable spatial anchors. 

For positions predicted at the current step, we set $C^{(t)}_j=c_j$. The local consensus around position $i$ is computed by fixed-window averaging:
\begin{equation}
\bar{c}_{\mathcal{N}_i}
=
\frac{1}{|\mathcal{N}_i|}
\sum_{j\in\mathcal{N}_i} C^{(t)}_j ,
\end{equation}
where $\mathcal{N}_i$ denotes a local window centered at position $i$.

We then refine the confidence score using a reliability-gated update:
\begin{equation}
\tilde{c}_i =
\begin{cases}
(1-\lambda)c_i + \lambda \bar{c}_{\mathcal{N}_i}, & c_i < \tau,\\
c_i, & c_i \geq \tau.
\end{cases}
\end{equation}
High-confidence predictions are kept unchanged, while uncertain predictions are re-ranked according to their agreement with local spatial anchors. SCR therefore improves spatial consistency at the token-ranking level without directly smoothing image pixels or modifying finalized token values.

\begin{figure*}[!t]
  \centering
  \includegraphics[width=\linewidth]{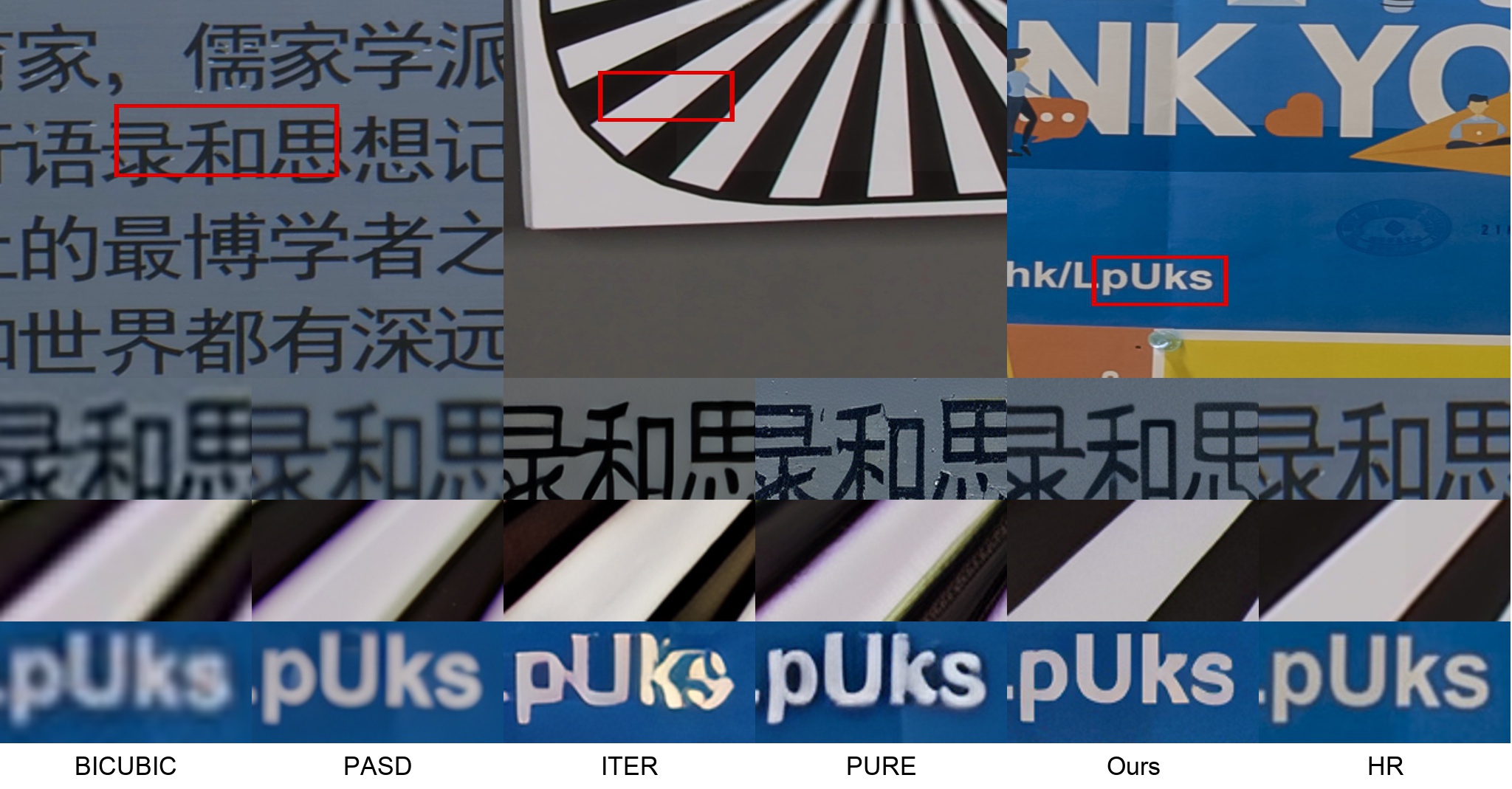} 
  \caption{Qualitative comparison for $\times4$ super-resolution. Compared with PASD and token-based baselines, DiMOO-SR produces sharper local structures, more legible text, and fewer isolated artifacts.}
  \label{fig:qualitative}
\end{figure*}

\section{Experiments}
\subsection{Experimental Settings}

\noindent\textbf{Datasets.} We evaluate DiMOO-SR on two widely used real-world super-resolution benchmarks: RealSR~\cite{cai2019toward} and DRealSR~\cite{wei2020component}. For training, we construct a dataset of approximately 310,000 high-texture patches from the first 10,000 images of FFHQ~\cite{karras2019style} and LSDIR~\cite{li2023lsdir}. Each training sample consists of an HR-LR-caption triple, where LR images are synthesized using the Real-ESRGAN~\cite{wang2021real} degradation pipeline, and detailed captions are generated by Qwen3-VL. For evaluation, we use fixed $512 \times 512$ patches sampled from the RealSR and DRealSR test sets. All compared methods are evaluated on the same patch list, rather than method-specific random crops, to ensure reproducibility and fair comparison.

\noindent\textbf{Implementation Details.} We fine-tune the Lumina-DiMOO backbone, while the VQ-GAN and text encoders remain frozen. Training uses AdamW with learning rate $5\times10^{-5}$ and cosine decay. The total batch size is 32 across 4 NVIDIA H100 GPUs. For Inverse Frequency Sampling (IFS), we set the temperature $\alpha=0.4$. During inference, SCR uses a $5\times5$ neighborhood, fusion weight $\lambda=0.7$, and trust threshold $\tau=0.5$. We employ $T=5$ sampling steps for all main experiments.

\subsection{Comparison with Representative Methods}
\label{subsec:comparison}

We benchmark DiMOO-SR against representative continuous diffusion and token-based generative SR methods. Our evaluation includes continuous diffusion models such as DiffBIR~\cite{lin2024diffbir}, PASD~\cite{yang2024pixel}, SeeSR~\cite{wu2024seesr}, and recent token-based generative models such as PURE~\cite{Wei_2025_ICCV}, ITER~\cite{chen2024iter}, Lumina-DiMOO~\cite{xin2025lumina}. Quantitative comparisons on the RealSR and DRealSR benchmarks are presented in Table~\ref{tab:cmp_sota}.
\begin{figure}[!ht]
  \centering
  \begin{subfigure}[b]{0.51\linewidth}
    \centering
    \includegraphics[width=\linewidth]{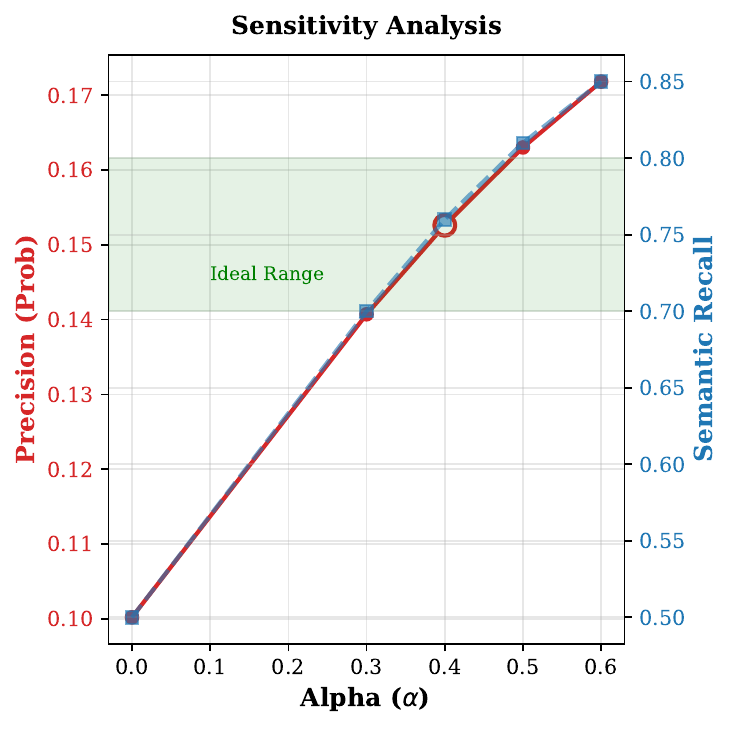}
    \caption{Sensitivity}
    \label{fig:diag_trend}
  \end{subfigure}
  \hspace{-2mm}
  \begin{subfigure}[b]{0.44\linewidth}
    \centering
    \includegraphics[width=0.9\linewidth]{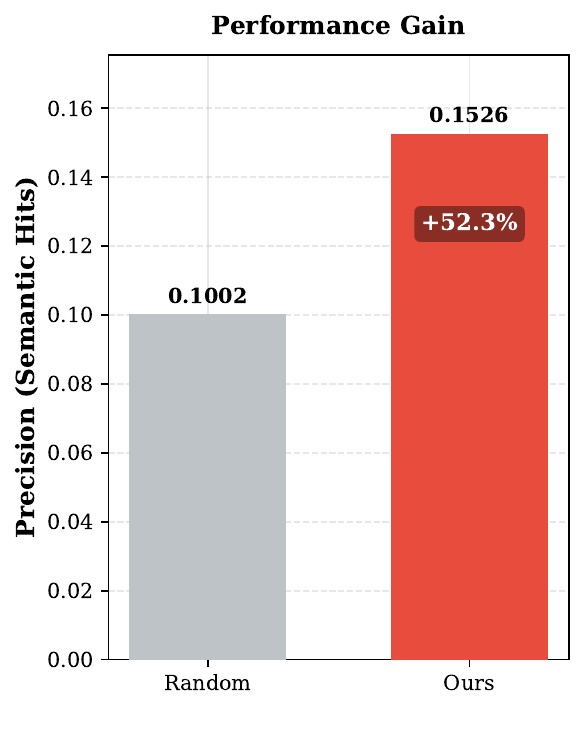}
    \caption{Gain}
    \label{fig:diag_bar}
  \end{subfigure}

  \vspace{2mm}
  \begin{subfigure}[b]{0.95\linewidth}
    \centering
    \includegraphics[width=\linewidth]{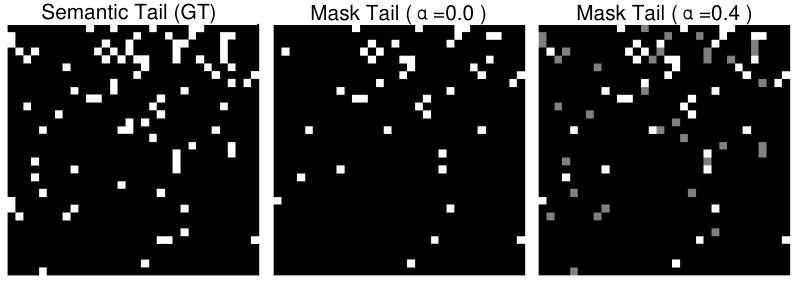}
    \caption{Visualization}
    \label{fig:diag_vis}
  \end{subfigure}
  \caption{Masking strategy diagnostics. Sweeping $\alpha$ reveals a precision--recall trade-off. Our default $\alpha=0.4$ focuses over 50\% more on semantic details compared with random masking, while the visualization shows that IFS selects rare texture and structure tokens.}
  \label{fig:diagnostic_results}
\end{figure}

\noindent\textbf{Quantitative results.} 
Table~\ref{tab:cmp_sota} compares DiMOO-SR with representative continuous diffusion and token-based generative SR methods.
Compared with continuous diffusion baselines, DiMOO-SR achieves favorable distributional and reference-based perceptual results.
On RealSR, DiMOO-SR obtains the lowest FID of 123.78 and DISTS of 0.2110, improving over SeeSR by 1.77 FID and 0.0113 DISTS.
On DRealSR, DiMOO-SR also reduces FID from 147.39 to 143.45 and DISTS from 0.2315 to 0.2259 compared with SeeSR.
These results indicate that discrete diffusion can provide competitive perceptual and distributional quality compared with continuous diffusion methods.

Compared with token-based generative models, DiMOO-SR brings clearer improvements.
On RealSR, it reduces FID from 133.54 to 123.78 and DISTS from 0.2233 to 0.2110 compared with Lumina-DiMOO.
On DRealSR, it reduces FID from 153.56 to 143.45 and DISTS from 0.2383 to 0.2259.
These gains suggest that rarity-aware training and spatially consistent decoding are important for adapting large multimodal discrete diffusion models to perception-oriented SR.

\noindent\textbf{Qualitative results.}
To further evaluate the perceptual quality of generated images, we present visual comparisons across challenging real-world scenarios in Figure~\ref{fig:qualitative}. First, in text reconstruction (top row), Bicubic and PASD leave the characters blurred, while ITER and PURE introduce unstable stroke shapes or local artifacts. DiMOO-SR recovers clearer character boundaries with faithful intensity and structure. Second, in geometric reconstruction (middle row), baselines like PASD and PURE introduce noticeable chromatic aliasing (purple/green fringing) and blurring along the diagonal lines. In contrast, DiMOO-SR accurately restores the sharp, monochromatic edges without color artifacts. Third, for small sign text (bottom row), ITER generates malformed letters and continuous diffusion baselines remain blurry. DiMOO-SR produces more legible text and cleaner local boundaries.

\begin{table*}[!ht]
  \centering
  \caption{Ablation studies on RealSR ($\times4$). IFS improves rare-detail learning, SCR improves spatial coherence, and dense prompts provide the strongest semantic guidance.}
  \label{tab:ablation_main}
  \scriptsize
  \setlength{\tabcolsep}{2.2pt}
  \begin{subtable}[t]{0.32\textwidth}
    \centering
    \caption{Component analysis.}
    \begin{tabular}{lcccc}
      \toprule
      Config. & MUSIQ$\uparrow$ & FID$\downarrow$ & DISTS$\downarrow$ & NIQE$\downarrow$\\
      \midrule
      Lumina-DiMOO  & 58.96 & 133.54 & 0.2233 & 5.9649\\
      + IFS  & 62.57 & 127.60 & 0.2175 & 5.9703\\
      + SCR  & 59.60 & 130.43 & 0.2195 & 5.4027\\
      \rowcolor{red!10}\textbf{Ours} & \textbf{63.76} & \textbf{123.78} & \textbf{0.2110} & \textbf{5.3636}\\
      \bottomrule
    \end{tabular}
  \end{subtable}
  \hfill
  \begin{subtable}[t]{0.32\textwidth}
    \centering
    \setlength{\tabcolsep}{3.8pt}
    \caption{SCR design.}
    \begin{tabular}{lcccc}
      \toprule
      Method & LPIPS$\downarrow$ & DISTS$\downarrow$ & MUSIQ$\uparrow$ & PSNR$\uparrow$\\
      \midrule
      Median & 0.3099 & 0.2139 & 63.55 & 24.07\\
      Gaussian & 0.3098 & 0.2130 & 63.60 & 24.12\\
      \rowcolor{red!10}\textbf{SCR} & \textbf{0.3053} & \textbf{0.2110} & \textbf{63.76} & \textbf{24.21}\\
      \bottomrule
    \end{tabular}
  \end{subtable}
  \hfill
  \begin{subtable}[t]{0.32\textwidth}
    \centering
    \setlength{\tabcolsep}{5.2pt}
    \caption{Prompt granularity.}
    \begin{tabular}{lccc}
      \toprule
      Prompt & FID$\downarrow$ & DISTS$\downarrow$ & NIQE$\downarrow$\\
      \midrule
      Null & 129.42 & 0.2166 & 5.3705\\
      Tags & 126.59 & 0.2175 & 5.4481\\
      \rowcolor{red!10}\textbf{Dense} & \textbf{123.78} & \textbf{0.2110} & \textbf{5.3636}\\
      \bottomrule
    \end{tabular}
  \end{subtable}
\end{table*}

\subsection{Analysis of IFS}
We evaluate the effectiveness of our masking strategy using two complementary metrics: \textit{Precision} (the proportion of masked tokens belonging to the semantic tail) and \textit{Semantic Recall} (the proportion of tail tokens successfully masked). As illustrated in Figure~\ref{fig:diag_trend}, varying the information temperature $\alpha$ reveals a distinct trade-off. At $\alpha=0$ (approximating uniform random masking), the model achieves only 50\% semantic recall, implying that half of the computational budget is expended on relearning trivial background patterns. As $\alpha$ increases, the masking distribution shifts towards rare tokens, improving both precision and recall. We identify $\alpha = 0.4$ as the optimal operating point. As quantified in Figure~\ref{fig:diag_bar}, this configuration achieves a precision of 0.1526, representing a substantial +52.3\% improvement over the random baseline, while maintaining a high semantic recall of 76.2\%. This balance ensures the model concentrates its learning capacity on challenging high-frequency regions without losing essential structural context. Figure~\ref{fig:diag_vis} visually corroborates this behavior: compared to random masking, our strategy ($\alpha=0.4$) selectively targets complex texture regions (visualized in Gray) to support stable generation.

\begin{figure}[!ht]
  \centering
  \includegraphics[width=\linewidth]{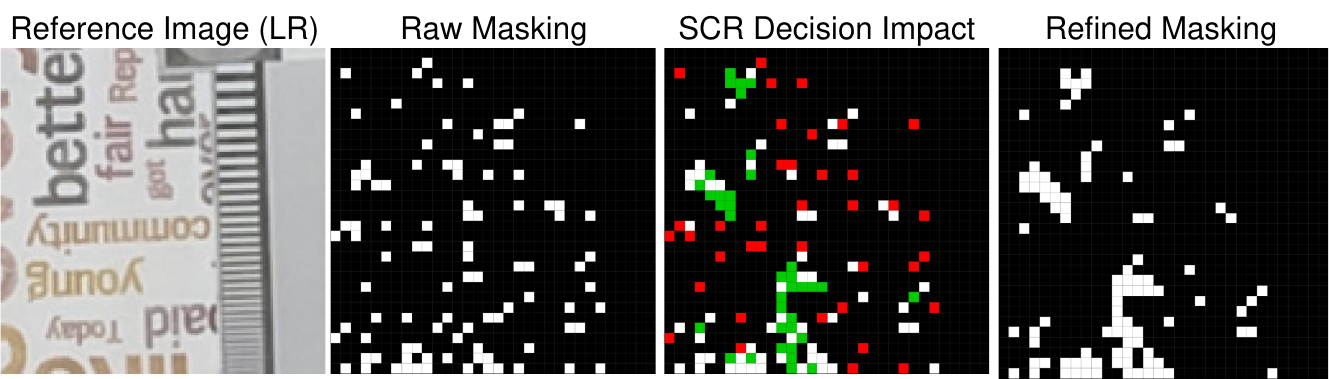}
  \caption{Visualizing the impact of SCR. The raw masking pattern contains scattered incoherent token decisions. SCR removes isolated artifacts and rescues coherent details using neighborhood consensus.}
  \label{fig:scr_vis}
\end{figure}
\begin{table}[!ht]
  \centering
  \caption{Complexity comparison among different methods. All methods are tested with an input
  image of size $512\times512$, and the inference time is measured on a NVIDIA 4090 GPU. Best are marked in Bold.}
  \label{tab:ablation_complexity}
  \resizebox{\linewidth}{!}{
  \begin{threeparttable}
  \begin{tabular}{lccccc}
    \toprule
    \textbf{Model} & \textbf{SeeSR} & \bf PASD & \textbf{PURE} & \textbf{Lumina-DiMOO} & \textbf{Ours} \\
    \midrule
    \# Params (M) & \bf 1900 & 2524 &  7080 & 8081 & \cellcolor{red!10} 8081 \\
    \# Time (s)   & 4.33 & 5.42 & 232.4 & 46.3 & \cellcolor{red!10} \bf 1.63 \\
    \bottomrule
  \end{tabular}
  \end{threeparttable}
  }
\end{table}

\subsection{Ablation Study}
\label{subsec:ablation}

\noindent\textbf{Effectiveness of Core Components}
Table~\ref{tab:ablation_main}(a) shows that IFS and SCR contribute complementary gains. IFS lowers FID from 133.54 to 127.60 and improves MUSIQ by exposing the model to rare structural and texture tokens more often during training. SCR reduces NIQE from 5.9649 to 5.4027 by suppressing spatially isolated low-confidence predictions at inference. Combining both modules gives the best overall perceptual balance, improving MUSIQ, FID, and NIQE.

\noindent\textbf{Why SCR instead of generic smoothing?}
Figure~\ref{fig:scr_vis} illustrates how SCR refines the masking decision. The raw masking pattern contains scattered uncertain token decisions, while SCR delays unreliable isolated predictions and retains locally coherent details through anchor-aware local consensus. We further compare SCR with median and Gaussian confidence filtering in Table~\ref{tab:ablation_main}(b). Unlike generic filters that uniformly process the confidence field, SCR is reliability-gated: retained tokens serve as spatial anchors, high-confidence predictions remain unchanged, and only uncertain predictions are re-ranked. As shown in Table~\ref{tab:ablation_main}(b), SCR achieves better LPIPS, DISTS, MUSIQ, and PSNR, demonstrating the effectiveness of the proposed ranking strategy.

\begin{figure}[!ht]
  \centering
  \includegraphics[width=.92\linewidth]{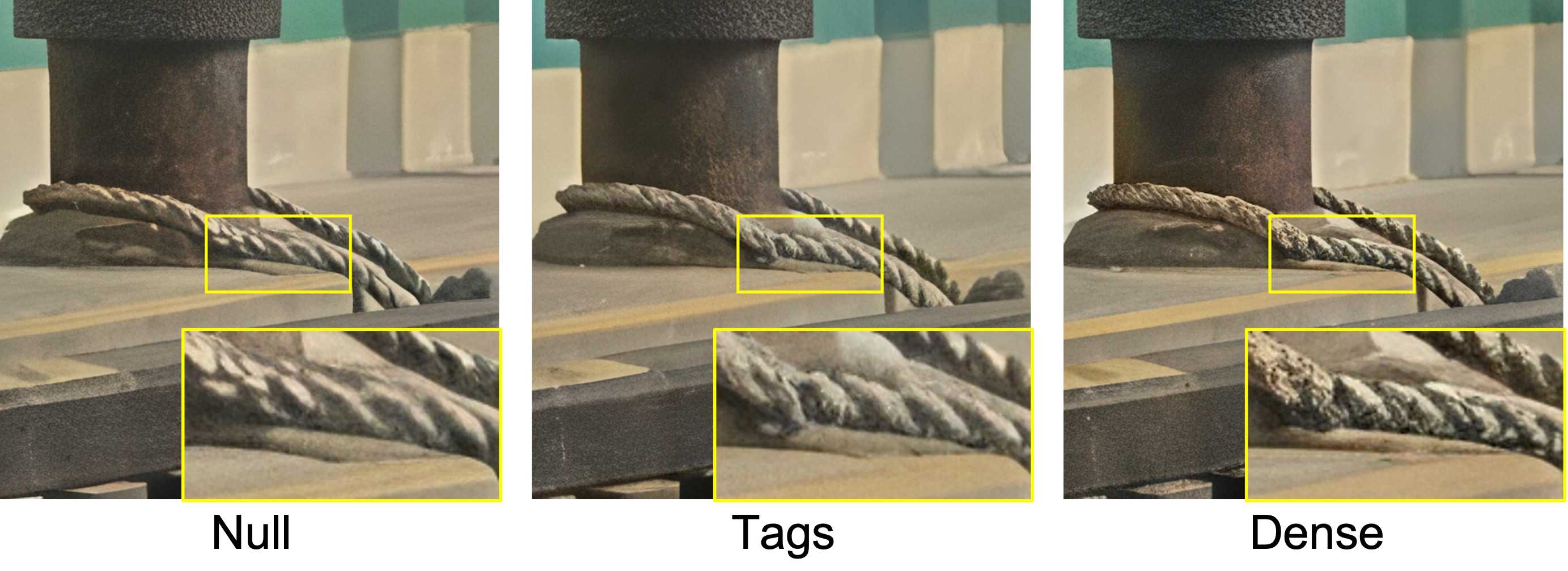}
  \caption{Subjective comparison of different text granularities. Dense captions provide fine texture, enabling more semantically faithful reconstruction than empty or coarse tag prompts.}
  \label{fig:text_finegrain}
\end{figure}

\noindent\textbf{Prompt granularity.}
As shown in Table~\ref{tab:ablation_main}(c), dense captions obtain the best FID and NIQE, indicating that fine-grained textual cues are useful even for a non-causal DDM decoder. Empty prompts force the model to rely mostly on LR visual evidence, while coarse tags supply useful but less localized hints. Dense captions include material, texture, layout, and local attribute descriptions, helping the model recover sharper text and more plausible fine structures. Figure~\ref{fig:text_finegrain} provides the subjective comparison, supporting the quantitative trend.

\noindent\textbf{Efficiency Analysis.}
Table~\ref{tab:ablation_complexity} compares the model scale and inference latency of representative generative SR methods. Although DiMOO-SR adopts a large multimodal backbone, it restores a $512\times512$ image in 1.63 seconds with 5 parallel decoding steps. Compared with PURE, DiMOO-SR avoids token-by-token autoregressive decoding. Compared with Lumina-DiMOO, the SR-specific decoding design with only 5 refinement steps substantially reduces latency. This shows that discrete diffusion provides a practical way to retain token-based multimodal generation while mitigating the sequential decoding bottleneck of autoregressive SR. More detailed efficiency analysis and additional comparisons are provided in the supplementary material.

\section{Conclusion}
\label{sec:conclusion}

In this work, we presented \textbf{DiMOO-SR}, a rarity-aware discrete diffusion framework for photo-realistic image super-resolution. We identified two practical challenges when adapting discrete diffusion to SR: the long-tailed distribution of visual tokens during training and spatial inconsistency during parallel decoding. To address these issues, we introduced Inverse Frequency Sampling (IFS) to reallocate supervision toward rare but information-rich tokens, and Spatial Consistency Ranking (SCR) to refine token confidence using local neighborhood agreement. Experiments on RealSR and DRealSR demonstrate that DiMOO-SR improves perception-oriented and distributional metrics, suggesting that discrete diffusion is a promising paradigm for generative image super-resolution.

\textbf{Limitations and future work.} DiMOO-SR inherits the granularity limitation of discrete token representations, which may restrict the recovery of extremely fine textures beyond the capacity of the underlying codebook. In addition, the use of an external VLM for dense caption generation introduces modular preprocessing latency. Future work may explore adaptive or hierarchical codebooks, lightweight captioning modules, and more effective fidelity control for discrete generative SR.

\bibliography{paperrefs}
\end{document}